# Retrieval-Augmented Clinical Benchmarking for Contextual Model Testing in Kenyan Primary Care: A Methodology Paper

Authors: Fred Mutisya (MBChB)1,2, Shikoh Gitau(PhD)1, Christine Syovata (MBChB)2, Diana Oigara (MBChB)2, Ibrahim Matende (MMED)2, Muna Aden( MBChB)2, Munira Ali (MD)2, Ryan Nyotu (MBChB)2, Diana Marion(MBChB, MBA)2, Job Nyangena( MBChB)2, Nasubo Ongoma1, Keith Mbae1, Elizabeth Wamicha(PhD)1, Eric Mibuari (PhD)1, Jean Philbert Nsengemana(MSc, MBA, MPA)3, Talkmore Chidede (PhD)4

Affiliation 1. Qhala,  2. Kenya Medical Association,  3. Africa CDC  4. AfCFTA

## Abstract

Large Language Models (LLMs) hold promise for improving healthcare access in low-resource settings, but their effectiveness in African primary care contexts remains under-explored. We present a rigorous methodology for creating a benchmark dataset and evaluation framework focused on Kenyan Level 2–3 (dispensary and health center) clinical care. Our approach leverages retrieval-augmented generation (RAG) to ground questions and answers in Kenya's national clinical guidelines, ensuring content aligns with local standard-of-care. The guidelines were digitised, chunked, and indexed for efficient semantic retrieval. Gemini Flash 2.0 Lite was then prompted with relevant guideline excerpts to generate realistic clinical questions, multiple-choice answers, and reasoning scenarios with source citations in English and Swahili. We engaged Kenyan physicians in a co-creation process to refine the dataset's relevance and fairness, and instituted a blinded expert validation pipeline to review for clinical accuracy, clarity, and cultural appropriateness. The resulting Alama Health QA dataset comprises thousands of regulator-aligned question-answer pairs spanning common outpatient conditions in English and Swahili. Beyond standard accuracy metrics, we propose innovative evaluation measures targeting clinical reasoning, safety, and adaptability (e.g. "Needle-in-the-Haystack" rare-case detection, Decision-Points stepwise reasoning, Geographic-Contextual Adaptability). Initial results highlight significant performance gaps in state-of-the-art LLMs when confronted with localized scenarios, echoing recent findings that LLM accuracy on African medical questions lags behind performance on U.S. benchmarks. Our work demonstrates a pathway for dynamic, locally-grounded benchmarks that can evolve with guidelines, providing a crucial tool for safe and effective deployment of AI in African healthcare.

# 1. Introduction

Advances in large language models have spurred interest in their potential to augment medical services, especially in low- and middle-income countries facing clinician shortages(Bekbolatova et al., 2024). By handling routine queries or providing decision support, LLMs might help bridge gaps in healthcare access across Africa. However, most existing medical AI benchmarks – such as USMLE-style question banks – reflect Western training and standards(Pal et al., 2022). Their content may not align with African epidemiology, available resources, or national treatment protocols. Indeed, a recent Pan-African evaluation (AfriMed-QA) found that while top LLMs can pass U.S. medical exams, their accuracy on African medical school questions is significantly lower, with performance varying widely by specialty and geography(Olatunji et al., 2024). This underscores the need for localized benchmarks that test models on the knowledge and reasoning required in African contexts.

**Why guidelines as the knowledge source?**

Health guidelines are best-practice statements regarding screening, diagnosis, management or monitoring of different health conditions(Kredo et al., 2016). They are the source of quality of care standards(National Institute for Care and Health Excellence, 2025). Health care is highly variable with each patient encounter being unique. This variability in care has resulted in the use of guidelines to standardize care(Aprikian, 2023). The world health organization has over 400 clinical practice guidelines on their portal(WHO, 2025). In Kenya and many countries, ministries of health issue clinical practice guidelines to standardize care. A survey of clinical practice guidelines in Kenya found 95 unique clinical care guidelines in use(Sagam et al., 2023).These documents distill evidence-based recommendations tailored to local realities (e.g. disease prevalence, resource availability) and are considered authoritative references for front-line providers. When applied correctly, such guideline recommendations "save lives" by ensuring consistent, quality care(WHO, 2025). We therefore base our benchmark on the Kenyan national guidelines for primary care (Level 2–3) (Kenya Ministry of Health, 2024)– as opposed to using ad-hoc question banks or foreign curricula. This approach grounds the benchmark in regulator-approved standards, making it directly relevant to real-world practice and safety. For example, Kenya's guideline for pneumonia in children specifies that severe pneumonia should be treated with benzyl penicillin plus gentamicin. A question derived from this guideline ensures an LLM must know and follow that local standard, rather than, say, a regimen from a different country. By constructing our QA items from such guidelines, we not only capture essential factual knowledge (first-line treatments, referral indications, etc.) but also encode context-specific norms (e.g. which medications are available at dispensaries, or what tasks a nurse versus a doctor should do at that level of care). This stands in contrast to generic medical Q&A datasets that may omit important local factors or use outdated information.

Benchmark overview: We introduce a pipeline for creating quantitative and qualitative benchmarks explicitly linked clinical practice guidelines. For this proof of concept English and Swahili language quantitative benchmarks of the Kenyan clinical guidelines for level 2 and 3 were created. This is detailed in the companion publication comparing this with other quantitative fact based benchmarks. The pipeline leverages Retrieval-Augmented Generation (RAG) methods (Lewis et al., 2021) that combine LLMs with document retrieval. RAG offers a pragmatic solution to LLM knowledge gaps and hallucinations by injecting up-to-date external information (i.e. the guideline text) into the prompt, thus constraining the model's output to be factual and source-backed. Recent surveys highlight that evaluating such hybrid systems requires assessing both the retrieval and generation components (e.g. document relevance, answer correctness, faithfulness to sources)(Huang & Huang, 2024). Accordingly, our work not only creates a dataset, but also proposes a comprehensive evaluation framework for LLM performance on it – moving beyond simple accuracy to also measure reasoning steps, contextual adaptability, and safety. We intentionally avoided replicating existing work in other benchmarks like HealthBench(Arora et al., 2024) to ensure that the methodologies being proposed were indeed novel. However, our process allows for customization of the pipeline to fit additional metrics of interest.

In the rest of this paper, we detail our methodology for knowledge base construction, question generation, and expert validation (Section 2). We then describe the structure and contents of the resulting benchmark dataset (Section 3). In Section 4, we introduce a set of innovative evaluation mechanisms designed to stress-test LLMs on clinical reasoning, drawing examples from the Kenyan & Swahili language primary care domain. Finally, we discuss implications, limitations, and future directions – including how our approach can be replicated for other countries and kept current with evolving guidelines. The aim is to provide a blueprint for building robust, context-aware benchmarks that can guide the safe deployment of AI in African healthcare.

## 2. Methodology

The methodology began with co-creation with healthcare workers to identify relevant clinical priorities, followed by the curation of clinical guidelines as the authoritative knowledge base. These guidelines were then processed through a retrieval pipeline that chunked and organized them for machine use. Next, a generation pipeline employed generative AI to create local-language question-answer datasets and reasoning questions based on the guidelines, and finally, the outputs underwent testing and validation with iterative refinements to ensure quality and contextual relevance (Figure 1).

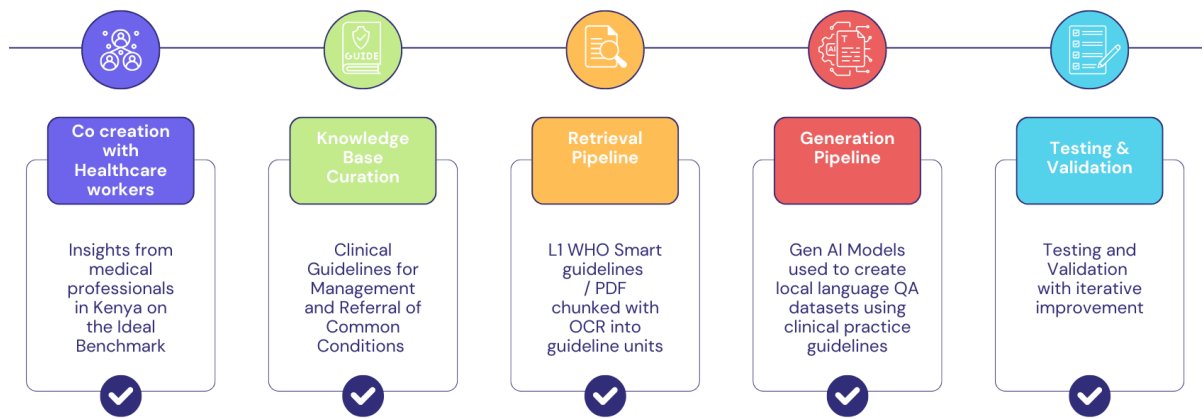

*Figure 1: Alama QA methodological pipeline*

## 2.1 Co-Creation with Domain Experts

A distinguishing aspect of our methodology is the co-creation process involving local medical experts. We used in-depth interviews and notes from the Qhala AI roundtable sessions with clinicians to define parameters. Examples of how medical expert feedback was converted to benchmark elements is shown in table 1

Table 1: Implementation of the Alama QA Co-Creation process

| Aspect | Co-creation Feedback | Our Implementation |
|---|---|---|
| Local Context Relevance | "I want each question to specify context – the patient's age, location, etc." | Added metadata or details (age, gender, setting) in scenarios to reflect time/place/person context. |
| Guideline Alignment | "Each answer should clearly state its source." | Every question's correct answer is explained with a guideline citation (source line or page) to ensure traceability. |
| Expert Validation | "We should involve more experts to vet the questions." | Developed a structured review rubric and expanded reviewer pool (beyond physicians to pharmacists and nurses) for diverse expert input. |
| Update Mechanism | "I am worried that guidelines change and some answers will become outdated." | Established pipeline tags for guideline version and an update schedule; flagged questions tied to potentially changing recommendations for periodic review. |

| Corpus Balance | "Ensure that no domain dominates" | Capped number of questions per specialty proportional to its content in guidelines (e.g. not too many HIV questions at expense of maternal health). |
|---|---|---|
| Topic Coverage | "Include emerging diseases and One Health topics." | Added content on newer priority areas (e.g. COVID-19, zoonoses) by incorporating recent WHO guidance; ensured coverage roughly aligns with disease burden evidence. |
| Question Diversity | "Don't make all questions too polished or similar." | Mixed question formats (MCQ vs. open-ended). Curated 5 unique |
| Multilingual Support | "Consider local languages for patient-facing questions." | Translated a subset (~10%) of the questions into Kiswahili, especially those involving patient advice. (All questions remain available in English as well for broader use.) |

Through this iterative process, we aligned the benchmark more closely with end-user expectations and ethical considerations. Notably, the co-creation helped us realize the importance of including contextual metadata: for instance, specifying "[Rural dispensary setting]" in a question stem if the scenario assumes limited resources, or indicating the time frame (e.g. postpartum period in a maternity case). These details ensure that when an LLM answers, it has to consider the context (something clinicians do intuitively).

Another example is multilingual support – an idea stemming from stakeholder comments that real patients might communicate in Kiswahili or local dialects. While our primary dataset is in English (for compatibility with model training and evaluation), we used the Gemini flash 2.0 lite model to generate Kiswahili versions, then had bilingual clinicians refine the translations.

## 2.2 Knowledge Base Construction

Our next step was to create a digital knowledge base of official clinical guidance. With close to 100 guidelines to choose from we selected the key primary care guidelines for this proof of concept. However, this can be substituted for any clinical practice guideline. We obtained the "Clinical Guidelines for Management and Referral of Common Conditions at Levels 2–3 (Primary Care)", published by the Kenya Ministry of Health (MoH) (Kenya Ministry of Health, 2024). This document (Volume 2 of the national guidelines series) encompasses outpatient diagnostics and treatments for a wide range of conditions seen at dispensaries and health centers – from infectious diseases like malaria and HIV, to maternal-child health, non-communicable illnesses, and basic emergency care. The guidelines are organized by body system and include defined referral criteria, aligning with Kenya's tiered health system.

The clinical practice guideline stretched across 636 pages and roughly 416 000 words, making it more a comprehensive primary-care textbook than a pocket manual. It was organized into nine substantive parts that mirrored the service-delivery realities of dispensaries and health centres (Table 2).

Table 2: Summary of Knowledge base

| Part | % of guideline | Top-level topics covered |
| --- | --- | --- |
| I. Internal Medicine & related | 26 % | Acute emergencies, HIV/STI, cardiovascular, CNS, endocrine, GI, infectious, rheumatology, renal, dermatology, mental health, adult pneumonia/TB |
| II. Paediatrics | 31 % | Paediatric resuscitation, diarrhoeal disease, malaria, meningitis, respiratory protocols, neonatology, immunization schedule |
| III. Surgery & related disciplines | 12 % | Trauma, burns, general surgery, dental/oral, ophthalmology, orthopaedics, ENT |
| IV. Obstetrics & Gynaecology | 10 % | Antenatal/intrapartum/post-partum care, common GYN disorders, contraception |
| V. Principles of Oxygen Therapy | < 1% | Indications, delivery methods, adverse effects |
| VI. Management of Blood & Blood Products | 2 % | Indications, paediatric & obstetric transfusion, reactions |
| VII. Referral Framework | 1% | Upward / downward referral algorithms, logistics, documentation |
| VIII. Forensic Medicine | 8 % | Legal foundations, injury documentation, mortuary practice |
| IX. COVID-19 annex | 3 % | Case definitions, IPC plan, paediatric vs adult severity tables & treatment pathways |

## Digitization and indexing

The first strategy in digitizing was to use the WHO smart-guideline L1 principle(Saban et al., 2024). This refers to guidelines in narrative format that have already been converted into a machine readable format. However, this is in the initial phases and is yet to be implemented for any guideline in Kenya. The alternative strategy relied on using the PDF versions of the guidelines.

We converted the PDF guideline into structured text and chunked it into semantically coherent sections (e.g. a paragraph and sub paragraphs). Each chunk was enriched with metadata (e.g. guideline source, section title, page) to facilitate retrieval and reverse compatibility. In total, 1115 guideline chunks were indexed. We had an existing hybrid retrieval strategy: (a) a BM25 lexical index for keyword matching, and (b) a vector similarity index for semantic search. However, since this first proof of concept QA was to create a dataset from the whole document, the retrieval step was skipped. This hybrid retrieval step ensures that a query can surface the most relevant guideline snippet even if it's phrased in layman terms or semantically related phrasing (for example, a query about "breastfeeding and HIV medication" would retrieve the PMTCT guideline section on antiretrovirals in lactating mothers).

## 2.2 Retrieval-Augmented Question Generation

We adopted a retrieval-augmented generation (RAG) pipeline to automatically produce question-answer items grounded in the above knowledge base. Figure 1 illustrates this pipeline. Each iteration proceeds as follows:

1. Retrieve a Guideline Snippet: We select a specific topic or subtopic from the guideline (e.g. "pneumonia in under-5s" or "diabetes management"). A retrieval query is issued to the knowledge base. For example, for the topic "childhood pneumonia," the system might retrieve the chunk of the guideline stating the criteria and first-line treatment for pneumonia in children under 5. For this proof of concept no specialty guideline was required hence all chunks were used without selective retrieval.

2. LLM Prompting to Generate Q&A: We feed the retrieved guideline excerpt to a Large Language Model with a carefully constructed prompt. The prompt instructs the LLM to "Use the provided guideline text to create a realistic question that tests understanding of this content. Then provide four answer options (A, B, C, D), mark the correct answer" An example prompt (in condensed form) is:

   *Prompt:*

   ```
   CASE_PROMPT = textwrap.dedent("""\
   ```

```
       You are a Kenyan healthcare expert who is a post-doctorate level exam
    constructor.
       Create {n} Kenyan healthcare training vignettes.  For each vignette:

         • 2-3-sentence scenario in a realistic Kenyan setting.
         • ONE MCQ in the exact format—

           Question: <question text>
           A) <option>
           B) <option>
           C) <option>
           D) <option>
           Correct: <letter>

       End every MCQ with "Correct:".

       Paragraph:
       \"\"\"{paragraph}\"\"\"
    """)
```

*Assistant (task output): Question:* "Q: According to Kenya's primary care guidelines, what is the first-line antibiotic regimen for a child with severe pneumonia?" *Options:* "A. High-dose amoxicillin oral for 5 days; B. Benzyl penicillin plus gentamicin; C. Chloramphenicol injection; D. Azithromycin plus ceftriaxone." *Answer:* "B. Benzyl penicillin plus gentamicin is correct. *Explanation:* The Kenyan Level 2–3 guidelines recommend benzyl penicillin together with gentamicin as the initial treatment for very severe pneumonia in children."

In producing the question, the LLM is forced to *stay true to the retrieved text*. The model effectively *"fills in the blank"* by turning a factual statement (e.g. *"First-line treatment is X"*) into a question (*"What is the first-line treatment?"*), and by inventing plausible distractors for the multiple-choice options. Importantly, because the model's knowledge is augmented with the exact guideline wording during generation, the chances of hallucinating an incorrect fact are minimized. The explanation and citation further enforce faithfulness – the model is prompted to explain the answer *using the guideline content*, which naturally leads it to paraphrase or quote the relevant lines (as shown in the example).

3. Model Selection for Generation: We initially experimented with three LLMs in the generation process: GPT-4 O mini and Gemini Flash 2.0 (lite) and LLaMA-3.1 (8B). All were run via API or local inference. We found that *Gemini Flash 2.0* produced the best balance of creativity and guideline adherence. Further testing is planned on the impact of varying the model types & parameters(e.g. temperature) on quality, cost and latency.

## 2.3 Expert Validation and Dataset Curation

After generation and initial filtering, all questions underwent a rigorous human validation process to ensure quality and safety. We collaborated with the Kenya Medical Association (KMA) and other healthcare professionals to review each question-answer pair. We built a web-based reviewing platform using Kobo Collect and adopted a blinded/masked review system: each question was independently evaluated by a licensed medical doctor, without knowing whether an item was from the Alama Health QA Pipeline or from 5 popular factual quantitative benchmarks. The reviewers used a standardized evaluation rubric with the following criteria (scored on a 5-point scale):

- Clinical Relevance: Is the question meaningful for a primary care setting in Kenya? (Does it reflect a real decision or important knowledge area?)

- Guideline Alignment: Is the stated correct answer actually correct per the cited guideline (and current standard of care)? Are any answer options or rationale contradicting the guidelines?

- Clarity and Completeness: Is the question phrased clearly, with all necessary information? (No ambiguous wording; if a scenario, it provides sufficient detail to answer.)

- Distractor Plausibility: Are the incorrect options plausible enough to be challenging (not too obviously wrong), yet clearly distinguishable by someone who knows the guideline?

- Language and Cultural Appropriateness: Is the wording culturally sensitive and free of bias or inappropriate assumptions? (E.g. using local names in cases, avoiding slang or potentially stigmatizing terms.)

Reviewers could also leave free-text comments to explain any issues or suggest improvements.

Dataset characteristics: Because our questions derive from an official compendium, they inherently focus on common and high-priority conditions. For example, there are dozens of questions on malaria, pneumonia, diarrheal disease, immunization, and maternal health – reflecting their weight in the primary care burden. The dataset is also dynamic: since each item is linked to a source, we can update or regenerate items if guidelines change (we plan to version the dataset alongside major MoH guideline revisions). This addresses a major limitation of static medical QA datasets noted by clinicians: they can become outdated as practices evolve. By maintaining a living knowledge base and an AI-assisted generation process, our benchmark can be refreshed with minimal manual effort – something not feasible with purely human-authored question banks. Finally, aligning with regulators (Ministry of Health and WHO) lends credibility:

an LLM that performs well on our benchmark would, in theory, be meeting the same standards expected of a human practitioner following national policy. This is an important consideration for model deployment and acceptance; regulators are more likely to endorse or approve AI tools that demonstrate adherence to official guidelines (since it correlates with patient safety and quality of care).

## 3. Evaluation Framework

Creating the benchmark dataset is only part of the contribution – we also devised a multi-faceted evaluation framework to *thoroughly assess LLM performance* on these clinical tasks. Traditional metrics like overall accuracy or F1 score on answers are useful but insufficient. We want to answer deeper questions: *Can the model reason through a case like a clinician? Does it catch critical cues? Will it adapt answers based on local context?* To this end, we propose a set of innovative evaluation mechanisms, inspired in part by clinical reasoning assessments and in part by the unique challenges of LLMs. Below, we describe each metric and provide an illustrative example drawn from the Kenyan primary care domain. These metrics can be applied as "challenge sets" or test scenarios in addition to scoring the model's answers on the Q&A dataset *(Figure 2)*.

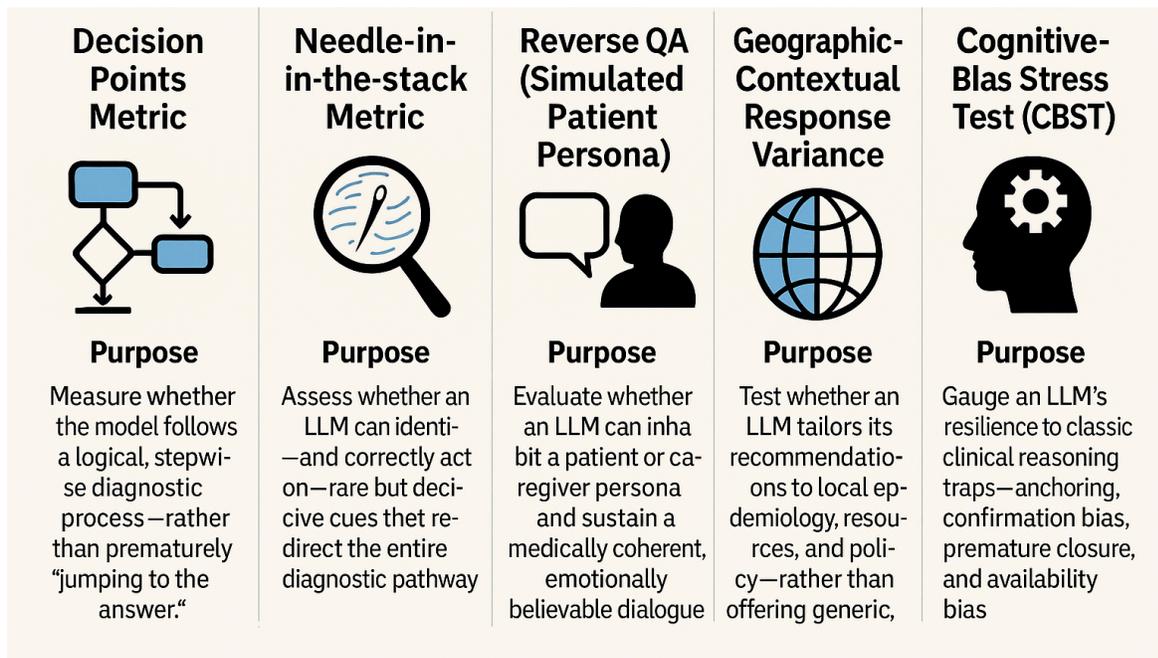

Figure 2: Reasoning model evaluation framework domains

### 3.1 Decision Points Metric

*Purpose*

Measure whether the model follows a logical, stepwise diagnostic process—rather than prematurely "jumping to the answer." We quantify how many clinically salient questions or actions (decision points) the model takes before issuing a final diagnosis or management plan.

*Existing Knowledge*

Existing health-LLM benchmarks such as HealthBench(Arora et al., 2024) and MedQA-USMLE(Singhal et al., 2023) include sequential reasoning items but do not explicitly count intermediate decision points. HealthBench's "Response depth" is the closest equivalent of this in existing benchmarks. Our metric builds on these foundations by (a) enforcing an interactive, clinician-like dialogue and (b) quantifying adherence to locally-relevant guideline branches, an element absent from prior benchmarks.

*Conventional Method*

We present the model with an open-ended case vignette and let it request additional information in an iterative, history-taking style dialogue.
Example vignette: *"A 5-year-old at a rural clinic presents with high fever and cough."*
At each turn we ask, "What would you like to ask or do next?" and supply the requested data until the model declares a diagnosis or treatment. We limit the token length to prevent information dumping.

A model that immediately states "This is pneumonia—treat with X" without seeking further history or examination receives a low score (premature closure). A model that systematically inquires about breathing rate, chest indrawing, oxygen saturation, prior vaccinations, and response to bronchodilators demonstrates structured reasoning and receives a higher score.

*Gen-AI Enhanced Method*

To scale and standardise curation of decision points across thousands of guideline-derived scenarios, we developed a retrieval-augmented pipeline:

1. Guideline Digitisation – High-resolution images of national algorithms (e.g., Kenyan IMNCI wall charts) are converted to structured text and flow-charts using Mistral-OCR, reducing manual entry time by >80 %.

2. Step Extraction – The Gen AI model parses the OCR output, identifies each decision node (symptom, sign, lab), and labels it with SNOMED or ICD-11 concepts.

3. Synthetic Case Generation – The Gen AI model receives the labelled flow-chart and produces diverse but guideline-consistent patient vignettes that require those decision nodes to be queried.

4. Evaluation Harness – During testing, the same flow-chart nodes are hidden from the candidate LLM. The harness logs whether and when the candidate queries each hidden node, enabling automated scoring at scale.

This Gen-AI pipeline lets us create thousands of high-fidelity, guideline-anchored scenarios across multiple disease areas in hours rather than weeks, while maintaining traceability back to the source algorithm.

*Scoring*

For each vignette we calculate:

- $N_{asked}$ – Number of guideline-critical nodes the model queried before its final answer.
- $N_{critical}$ – Total critical nodes for that vignette.
- Precision = $(N_{asked \cap critical}) / N_{asked}$
- Recall = $(N_{asked \cap critical}) / N_{critical}$

A composite Decision-Point Score = 2 × (Precision × Recall)/(Precision + Recall) (F-score) is rescaled to 0–10 for interpretability. Models that reach the correct diagnosis with high precision-recall—and without superfluous, non-guideline questions—receive the highest scores.

## 3.2 Needle-in-the-Haystack Metric

*Purpose*
Assess whether an LLM can identify—and correctly act on—rare but decisive cues that redirect the entire diagnostic pathway. The metric rewards vigilance and nuanced reading comprehension: can the model spot the "needle" (a subtle travel history, a single abnormal sign, an overlooked lab value) hidden in a voluminous clinical "haystack"?

*Existing Knowledge*
Benchmarks such as HealthBench(Arora et al., 2024), MedMCQA(Pal et al., 2022), and MultiMedQA(Singhal et al., 2023) test reading-comprehension and differential-generation but seldom isolate single game-changing clues. ARC-style "reading-comprehension" items embed key facts but lack clinical distractors, while USMLE vignettes often telegraph the diagnosis with multiple converging findings. Recent efforts like HealthBench introduced endemic-disease cases yet still evaluate only final-answer correctness. Our Needle-in-the-Haystack metric extends this lineage by (a) deliberately burying one decisive cue, (b) rewarding explicit recognition of that cue, and (c) rooting each cue in local epidemiology—addressing an unmet need for vigilance testing in guideline-reliant African settings.

*Conventional Method*
We create extended case vignettes packed with plausible yet distracting symptoms, demographic details, and lab results. One critical clue, often a single sentence or finding, is embedded mid-narrative.

*Example vignette*

A 32-year-old humanitarian worker presents with fever, myalgia, and abdominal pain. She reports progressive fatigue, mild jaundice, and weight loss. *She returned from Bentiu, South Sudan, four weeks ago after a six-month deployment.* Vitals, physical examination, and preliminary labs are provided in full.

A clinician-like LLM should ask clarifying questions, highlight the South Sudan travel, and elevate visceral leishmaniasis (Kala-azar) in its differential. A less vigilant model defaults to malaria or typhoid and never references the travel clue.

We grade the response as:

- 0 – clue ignored; incorrect diagnosis/plan.
- 0.5 – clue mentioned but not integrated into reasoning (lists Kala-azar but ranks it low and treats for malaria).
- 1 – clue explicitly used to justify a correct diagnosis or priority differential, with an appropriate management plan.

*Gen-AI Enhanced Method*

Manual crafting of long, distraction-rich vignettes is labour-intensive. We therefore built an automated, guideline-anchored pipeline to generate and validate "needle" cases at scale:

1. Clue Library Construction – Using Mistral-OCR, we convert national algorithms into structured text. A rule-based filter extracts *rare-but-critical* risk factors (e.g., "recent camel contact in Marsabit → CCHF", "untreated well water in Turkana → Guinea worm").

2. Synthetic Vignette Generator – The Gen AI model receives a template consisting of:
*Common symptom bundle + one "needle" clue + demographic setting + distractor comorbidities.*
It outputs coherent 300–500-word cases whose surface features obscure the underlying diagnosis unless the clue is recognized.

3. Automated Plausibility Audit – A verifier ensemble checks that:
    - the # of distractor diagnoses ≥ 3,
    - the "needle" clue is present exactly once,
    - guideline-consistent management exists for the target disease.

4. Scoring Harness – During evaluation, the harness scans the model's response for explicit mention of the clue or its downstream implication (using regex + embedding similarity). Partial-credit rules are applied automatically.

This pipeline produces hundreds of geographically-specific needle cases (Kala-azar in South Sudan returnees, Rift Valley fever in pastoralists, neurocysticercosis in pork-eating regions) in hours, ensuring diversity and reproducibility while preserving human-grade realism.

*Scoring*

For each vignette we compute:

| Variable | Definition |
|---|---|
| Clue Detected | Binary (1/0): model explicitly mentions or asks about the rare cue. |
| Correct Diagnosis | Binary: target disease included in top-3 differential *and* management aligns with guideline. |
| Needle Score | 1.0 if both *Clue Detected* and *Correct Diagnosis* are true; 0.5 if only one is true; 0 otherwise. |

Aggregate performance is reported as the mean Needle Score across all cases, with 95 % bootstrap CIs. High scores indicate the model consistently surfaces critical, low-frequency signals amid informational noise—mirroring expert clinical reasoning under real-world complexity.

### 3.3 Reverse QA (Simulated Patient Persona)

*Purpose*
Evaluate whether an LLM can *inhabit* a patient or caregiver persona and sustain a medically coherent, emotionally believable dialogue. The test flips the classic QA format: the clinician (human or AI agent) asks the questions, while the model answers as the patient or family member. Success indicates deep symptom understanding, narrative consistency, and appropriate tone—skills critical for virtual standardized-patient training and conversational triage systems.

*Existing Knowledge*
Existing patient simulator benchmarks include MedAgent Sim from Mohamed bin Zayed University of Artificial Intelligence(Almansoori et al., 2025). Reverse question answering is already an established LLM benchmarking task in the non medical field(Balepur et al., 2025). These prove feasibility but:

- rely on English-only personas (limiting cultural realism),
- use hand-written vignettes instead of guideline-derived fact sheets,
- grade primarily on fluency, not medical fidelity.

Our approach advances the field by anchoring every persona in *locally validated* algorithms (e.g. Kenyan Level 2/3 clinical guidelines, Ugandan IMNCI, South African EDL), layering dialectal nuance, and providing an automated contradiction detector—bridging the gap between global LLM evaluation and African bedside realities.

*Conventional Method*
1. Persona Fact-Sheet – A human curator writes a short, structured "ground-truth" card (e.g., *Child: 2 yrs, 5 watery stools/day, no urine for 12 h, sunken eyes, no vomiting*).

2. Prompt – *"You are the child's mother. Answer the doctor's questions truthfully and naturally, based only on the facts below…"*

3. Dialogue Session – A scripted clinician bot (or human evaluator) poses 6-10 standard questions about onset, hydration, danger signs, etc. The model replies in first-person.

4. Human Grading – Two clinicians score each response for (a) factual consistency with the card, (b) medical plausibility, and (c) style realism (concerned mother, appropriate Kenyan English + Swahili loanwords, no jargon).

Although reliable, the manual pipeline yields only dozens of cases per reviewer-day and suffers from inter-rater variability.

*Gen-AI Enhanced Method*

To generate thousands of culturally-anchored personas while safeguarding factual fidelity, we introduced a retrieval-augmented, multi-model stack:

| Stage | Function |
| --- | --- |
| 1. Guideline Parsing | Extract danger-sign tables, age-specific thresholds, and counselling scripts → JSON schema |
| 2. Persona Template Builder | Combines a random guideline chunk + demographic seed (age, caregiver role, county) → "fact sheet" |
| 3. Dialect & Emotion Layer | Infuses localized phrases ("hakuna mkojo", "amekosa nguvu kabisa") and caregiver affect tags (worried, hesitant, relieved) |
| 4. Conversational Simulator | Pairs the persona LLM with a scripted clinician bot; logs Q–A turns and auto-flags contradictions |
| 5. Automated Scoring | - *Consistency*: ≤ 1 contradiction across turns |

- *Completeness*: answers cover ≥ 90 % of fact-sheet items when asked

This pipeline converts raw guideline text into thousands of high-fidelity, language-appropriate simulated-patient dialogues in under a day, with machine-verified consistency.

*Scoring*

For each dialogue we compute:

| Metric | Formula (per case) | Weight |
| --- | --- | --- |
| Consistency Score | 1 – (# factual contradictions / # clinician questions) | 0.4 |
| Completeness | (# fact-sheet items correctly disclosed / total expected) | 0.3 |

| Style Realism | cosine-sim(persona embedding, reference emotion embedding) | 0.2 |
| Linguistic Appropriateness | binary: contains relevant local terms when prompted | 0.1 |

A weighted composite (0–10) constitutes the Reverse-QA Persona Score. Confidence intervals are boot-strapped over 1 000 cases. High-scoring models consistently convey accurate symptoms, maintain first-person tone, and mirror caregiver affect—yielding credible virtual patients for clinician training or benchmark stress-tests.

## 3.4 Geographic-Contextual Response Variance

*Purpose*
Test whether an LLM tailors its recommendations to local epidemiology, resources, and policy—rather than offering generic, "one-size-fits-all" advice. We use vaccination schedules as the exemplar domain because schedules differ sharply across countries (e.g., yellow-fever at 9 months in Kenya vs. none in the UK; hexavalent vs. pentavalent formulations; birth-dose Hep B availability). Correct answers *must* diverge by setting.

*Existing Knowledge*
Benchmarks like HealthBench(Arora et al., 2024), and MultiMedQA(Singhal et al., 2023) have no consistent geography variance with reliable metadata. Afrimed QA introduced endemic-disease diversity but still evaluates final-answer accuracy without penalising context insensitivity(Olatunji et al., 2024). The Vaccination based RAG AI-VaxGuide developed an automated question generation pipeline but was only based on the Algerian vaccination programs and the WHO schedule(Zeggai et al., 2025).Our metric advances the field by (1) pairing *identical* symptoms with divergent locales, (2) grounding every answer in up-to-date national schedules, and (3) explicitly grading resource-appropriate recommendations.

*Conventional Method*
1. Scenario Pairing – Craft two (or more) otherwise identical cases that vary only in geographic tag:
   *Case A*: "A 10-week-old infant in Johannesburg presents for routine immunisation."
   *Case B*: "A 10-week-old infant in Kisumu, Kenya presents for routine immunisation."

2. Prompt – *"Which vaccines are due today, and what counselling should be given?"*

3. Manual Review – Experts compare model outputs to the respective national EPI schedules. A correct Kenyan answer includes Pentavalent + PCV10 + OPV, whereas the South-African schedule gives Hexavalent + PCV13 + RV.*

Drawbacks: hand-authoring all pairs is labour-intensive, and reviewers must juggle dozens of country guidelines from memory.

*Gen-AI Enhanced Method*

| Stage | Function |
|---|---|
| 1. Multi-Guideline Knowledge Base | Scrape & digitise national immunisation schedules (WHO EPI Atlas, ministries of health PDFs). Store each dose as structured triples: *(country, age, antigen, formulation, constraints)*. |
| 2. Contextual Scenario Generator | Creates matched clinical vignettes that differ only by country, setting (urban tertiary hospital vs. rural dispensary), or stock-out flag. |
| 3. Answer Key Synthesiser | Pulls the correct vaccine list and feasible investigations *given resource tier* (e.g., no bilirubin test in Level 2 Kenyan facility). Generates ground-truth rationales. |
| 4. Evaluation Harness | Compares model response to ground-truth for (a) antigen set match, (b) formulation correctness, (c) acknowledgment of resource limits (cold-chain, stock-outs). Partial credit applied automatically. |

The pipeline yields hundreds of paired-context tasks spanning > 20 African and global immunisation programmes in under 12 hours, while preserving provenance links to each source document.

*Scoring*

For each scenario pair we derive four sub-scores:

| Component | Definition | Weight |
|---|---|---|
| Antigen Accuracy | Jaccard similarity between predicted and required vaccine set | 0.35 |
| Formulation Fit | Correct format (e.g., Pentavalent vs. Hexavalent) | 0.25 |
| Resource Alignment | Advice respects local availability / cold-chain limits | 0.25 |
| Rationale Localization | Explicit mention of locale-specific factors (endemic diseases, policy) | 0.15 |

The Context Adaptation Score (CAS) is the weighted sum, scaled 0–10.

A Delta-CAS across paired scenarios gauges flexibility: high-performing models change recommendations appropriately between Johannesburg and Kisumu, Lagos and London, or district hospital vs. dispensary. Models that recycle the same plan regardless of context receive low or even negative Δ-scores, signalling dangerous genericism.

## 3.5 Cognitive-Bias Stress Test (CBST)

*Purpose*

Gauge an LLM's resilience to classic clinical reasoning traps—anchoring, confirmation bias, premature closure, and availability bias. The metric asks: Will the model cling to its first impression, or will it appropriately revise its thinking when new, disconfirming evidence appears?

*Existing Knowledge*

Koo et al developed a cognitive bias benchmark called CoBBLEr which had various types of bias such as order bias where a model favours the order of responses rather than the quality and egocentric bias where its own responses are preferred (Koo et al., 2024). Other cognitive benchmarks include LM-EVAL HARNESS (Sutawika et al., 2025) and HELM (Liang et al., 2023). Our CBST bridges that gap by:

- grounding each bias trigger in evidence-based African guidelines,
- automating large-scale synthesis of anchor-and-flip scenarios, and
- scoring explicit revisions, not just final-answer accuracy.

Conventional Method

1. Bias-Priming Vignette – Craft a two-stage case:
   Stage 1 strongly implies a plausible but incorrect diagnosis (e.g., *"40-year-old man with retro-sternal burning; initial ECG normal → heartburn likely"*).
   Stage 2 adds contradictory or alarming data (persistent pain, risk factors, dynamic ECG changes).

2. Prompt Sequence –
   *"Given the information so far, what is your working diagnosis?"* → capture model's initial answer.
   Add Stage 2 details → *"Now update your diagnosis and plan."*

3. Human Review – Clinician graders check whether the model updates its differential, requests further tests, and de-anchors from the initial wrong lead.

Limitations: hand-writing bias cases is slow, and graders may subjectively judge what counts as *"adequate"* breadth.

Gen-AI Enhanced Method

| Stage | Component |
|---|---|
| 1. Bias-Template Library | Manual curation of ~20 archetypal bias scenarios (anchoring, premature closure, confirmation bias, etc.). |
| 2. Guideline-Linked Fact Injector | Mistral-OCR extract contradicting red-flag details from national algorithms (e.g., IMCI danger signs, STEMI criteria). |
| 3. Synthetic Case Engine | Gen AI model populates each template with random age, comorbidities, vitals, and *one* decisive late clue that flips the likely diagnosis. |
| 4. Ground-Truth Reasoning Chain | Gen AI generates the expected step-wise thought process: initial differential ➜ trigger to reconsider ➜ final correct diagnosis ➜ recommended tests/management. |
| 5. Auto-Scoring Harness | A BERT model compares the candidate LLM's *before* and *after* answers to the ground-truth reasoning chain: |

- Does the model mention contradicting clue?

- Does it broaden / change differential?

- Does it order confirmatory tests?

This pipeline produces hundreds of bias-primed scenarios across multiple specialties (cardiology, paediatrics, emergency) and resource tiers in < 8 hours, with machine-verifiable answer keys.

*Scoring*

| Metric | Definition | Weight |
|---|---|---|
| Anchor Flexibility | 1 if final diagnosis differs (appropriately) from the anchored first guess; 0 otherwise | 0.40 |
| Contradiction Recognition | Binary: model cites late "red-flag" detail when revising | 0.25 |
| Breadth of Differential | Normalised count of plausible alternatives listed after revision | 0.20 |
| Action Appropriateness | Orders or recommends guideline-aligned confirmatory tests / management | 0.15 |

Composite CBST Score = weighted sum (0–10).
We additionally report a Bias Susceptibility Index (BSI) = mean probability the model *fails* to revise when required. Lower BSI and higher CBST indicate robust, bias-aware reasoning—essential for safe deployment in varied African clinical contexts.

# 4. Discussion

Benchmark strengths: The Alama Health QA dataset and its associated evaluation framework offer a novel contribution at the intersection of *global AI technology and local healthcare knowledge*. By grounding content in Kenya's clinical guidelines, we ensure that the benchmark is clinically valid and locally relevant – a decided advantage over importing foreign medical exams. Our methodology demonstrates that retrieval-augmented generation can be harnessed to produce a high-quality, *evolving* question bank aligned with regulator expectations. This is in line with recent trends towards *dynamic benchmarks* that can keep pace with knowledge updates. We also showed the value of engaging local stakeholders (through co-creation and validation) to infuse context and ownership into the AI development process. This participatory approach likely improved the cultural appropriateness and acceptance of the final product. Furthermore, the comprehensive evaluation metrics we propose push the envelope beyond simple accuracy, touching on reasoning, ethics, and safety – aspects that are critical for deploying LLMs in real clinics, as highlighted by Elmitwalli *et al.* (2025) in their domain-specific evaluation framework.

Challenges and limitations: Despite its advantages, our approach has limitations. Firstly, the reliance on guidelines means the dataset's questions are as good as the guidelines themselves. If a guideline is outdated or has gaps, the generated questions might propagate those. Thus, it is recommended to use specialty guidelines instead of general purpose guidelines. On the evaluation side, one might argue our custom metrics introduce their own subjectivity – e.g., how to weigh the importance of each metric is somewhat arbitrary at this stage. We partly address this by not trying to collapse everything into a single score; instead, we envision a radar chart or profile for each model across metrics. Still, more formal reproducibility of these test scenarios is needed. We plan to open-source the evaluation scripts and a small "benchmark subset" of the dataset so that others can run the same tests, and we welcome feedback to refine the scoring rubrics.

Comparison with AfriMed-QA and other datasets: AfriMed-QA (Olatunji *et al.*, 2025) was a trailblazer in creating an African medical QA set. It drew on exam questions from many countries, resulting in a broad mix of content. Our work differs in that it is narrower in scope but deeper in alignment. AfriMed-QA covers 32 specialties with questions of varying difficulty (including very advanced ones from specialist exams) – it's excellent for testing broad knowledge. However, it may contain questions that assume tertiary-hospital contexts or outdated practice in some regions. In contrast, our dataset zeroes in on primary care and ensures every item is tied to a current guideline recommendation. One could say AfriMed-QA asks "What do African doctors collectively know or find important?", whereas Alama Health QA asks "What should a competent primary care provider in Kenya do according to policy?". Both are valuable: indeed, they could be used complementarily. An ideal LLM for African health should probably be trained and tested on a union of both – to both satisfy official guidelines (our focus) and answer the kind of ad-hoc questions front-line clinicians think of (the focus of AfriMed-QA's expert-derived questions).

Generality and future work: We have leveraged this proof of concept work to formally produce the first HIV benchmark in English and Swahili. The development is ongoing and will be open sourced when completed. We will also formally test both frontier models and small language models in use in Africa and present the results.

Implications: The development of this benchmark has practical implications for Kenya and beyond. For policymakers, having an evaluation dataset tied to their guidelines provides a tool to audit AI systems before they are approved for clinical use. We could imagine the Ministry of Health using our benchmark as a required "exam" for any AI health assistant entering the market – much like a board exam for doctors. If an AI can't score, say, above a certain threshold on questions derived from the MoH's own guidelines, it should not be trusted to advise on patient care in that jurisdiction. On the other hand, if it performs well, that could accelerate regulatory acceptance. There is also educational value: our question set, with explanations and guideline references, can serve as a study aid for medical trainees (essentially a comprehensive quiz on primary care).

Finally, we note that our project embodies a paradigm shift in benchmarking: moving away from static, competition-style leaderboards toward *community-driven, iterative, and context-rich benchmarks*. The involvement of local experts at every stage (content creation, validation, metric design) is resource-intensive but yields a product that is meaningful on the ground. We encourage AI researchers to adopt similar co-creation approaches, especially for applications in the Global South where context is everything. The success of AI in healthcare will depend not just on clever models, but on earning the trust of healthcare providers and patients – and that trust comes from demonstrating alignment with the standards those providers and patients expect. In summary, our work provides a template for ensuring "AI that *adheres to guidelines* is the AI that can be safely deployed," and offers a rich benchmark to measure progress toward that goal.

## 5. Conclusion

We have presented a comprehensive methodology for building and evaluating a retrieval-augmented LLM benchmark grounded in African healthcare guidelines. Using Kenya's primary care setting as a use-case, we demonstrated how to turn official clinical guidelines into a dynamic QA dataset, leveraging both state-of-the-art AI and local human expertise. The resulting benchmark not only tests factual knowledge but also probes an AI model's reasoning process, ethical judgment, adaptability to context, and commitment to patient safety – facets that are critical for real-world clinical utility. Our approach aligns AI behavior with standard-of-care guidelines, offering a clear path for regulators and practitioners to assess if a model is fit for deployment. While challenges remain, this work lays important groundwork toward trustworthy and locally relevant medical AI. We envision a future where each country or region can maintain its own AI benchmark tied to its health guidelines, and where LLMs aspiring to assist in healthcare must prove themselves on these benchmarks. Such an ecosystem would drive AI to truly "do no harm" and to be an asset rather than a liability in healthcare systems worldwide.

Future work will extend this paradigm to other African nations and clinical domains, fostering a network of benchmarks that collectively raise the standard for AI in global health. In doing so, we hope to catalyze a shift from AI models that are merely impressive in controlled settings, to AI systems that are *reliable partners on the front-lines* – supporting health workers, educating patients, and ultimately improving outcomes in the communities that need it most.